\documentclass[sigconf,nonacm]{acmart}

\setcopyright{none}
\usepackage{multirow}
\usepackage{microtype}
\usepackage{paralist}
\usepackage{svg}
\usepackage{paralist}
\usepackage{float}
\usepackage{rotating}
\usepackage{stfloats}
\pdfoutput=1

\usepackage[appendix=append]{apxproof}

\usepackage{newunicodechar}
\newunicodechar{γ}{$\gamma$}
\newunicodechar{⊆}{$\subseteq$}

\usepackage{booktabs}
\usepackage{array}

\author{Shandong Yuan}
\email{yuanshandong13@nudt.edu.cn}
\orcid{0000-0003-3433-0940}
\affiliation{%
	\institution{National University of Defense Technology}
	\city{Changsha}
	\country{China}
}

\author{Jun He}
\email{93545322@qq.com}
\orcid{}
\affiliation{%
	\institution{National University of Defense Technology}
	\city{Changsha}
	\country{China}
}

\begin{document}

\title{SECNN: Squeeze-and-Excitation Convolutional Neural Network for Sentence Classification}

\author{}

\begin{abstract}
Sentence classification is one of the basic tasks of natural language processing. Convolution neural network (CNN) has the ability to extract $n$-grams features through convolutional filters and capture local correlations between consecutive words in parallel, so CNN is a popular neural network architecture to dealing with the task. But restricted by the width of convolutional filters, it is difficult for CNN to capture long term contextual dependencies. Attention is a mechanism that considers global information and pays more attention to keywords in sentences, thus attention mechanism is cooperated with CNN network to improve performance in sentence classification task. In our work, we don’t focus on keyword in a sentence, but on which CNN’s output feature map is more important. We propose a Squeeze-and-Excitation Convolutional neural Network (SECNN) for sentence classification. SECNN takes the feature maps from multiple CNN as different channels of sentence representation, and then, we can utilize channel attention mechanism, that is SE attention mechanism, to enable the model to learn the attention weights of different channel features. The results show that our model achieves advanced performance in the sentence classification task.
\end{abstract}

\maketitle

\section{Introduction}
Sentence classification is to predict the label of each natural language statement by training a classifier. In natural language processing (NLP) domain, sentence classification is also a foundational component in other applications, such as sentiment analysis (SA), topic labeling (TL), question answering (QA), natural language inference (NLI), etc. CNN encodes $n$-grams by convolution operation and generates a fixed-size high-level representation, for sentence modeling, it is adept in extracting abstract and robust features of the input. The first introduction of CNN into sentence classification task is the work done by Kim\cite{Kim2014}. In recent years, CNN has become popular in NLP\cite{Kim2014,Kalchbrenner2014}, and has achieved remarkable achievements in sentence classification\cite{Kim2014,Kalchbrenner2014,Zhang2015,Zhou2015,Johnson2014,Young2016}. However, there is a limitation of CNN that it only considers sequential $n$-grams, and thus ignores some long-distance correlations between non-consecutive words, which play an important role in many linguistic phenomena.

Attention is a mechanism which can automatically capture long-term contextual information and correlation between non-consecutive words without any external syntactic information. Therefore, recently, the attention mechanism cooperates with CNN to deal with NLP tasks. The first work that incorporates attention mechanism into CNN is completed by Yin\cite{Yin2016}, they proposed a novel model named Attention-Based BCNN (ABCNN) to model a pair of sentences, and presented three different attention mechanism based on Basic Bi-CNN (BCNN), Among them, ABCNN-1 calculates the attention weights directly on the input representation in order to improve the features computed by convolution; instead, ABCNN-2 calculates the attention weights on the output of convolution in order to reweight the convolution output; and ABCNN-3 combines ABCNN-1 and ABCNN-2 by stacking them. Zhao\cite{Zhao2016} proposed ATT-CNN model in which attention is introduced between input layer and convolution layer, the attention layer generates a context vector which will be concatenated with the word embedding as a new word representation. Lin\cite{Lin2017} introduced a self-attention mechanism into their model aiming at extracting sentence embedding, the attention mechanism enables the sentence to focus on specific elements of a sentence. Sun\cite{Sun2019} proposed a novel Inatt-MCNN model, which combines Inner-attention mechanism and multi-channel CNN, the Inner-attention mechanism aims to capture the key part of a sentence. Li\cite{Li2020} presented Adaptive Gate Attention model with Global Information (AGA+GI), in which the AGA integrates statistics feature into semantic features and the GI describes a global statistics of a term towards the labels. Zhang\cite{Zhang2022} designed a novel label-based attention mechanism to hierarchically extract important features of each document based on individual labels at each hierarchical level. Liu\cite{Liu2020} proposed Attention-based Multichannel CNN (AMCNN), in which they introduced two attention mechanisms to obtain multichannel representations, the first mechanism is scalar attention aiming at calculating the weights of input elements (i.e. word-level importance), and the second one is vectorial attention for calculating the weights of each dimension in the input element (i.e. feature-level importance). Kulkarni\cite{Kulkarni2021} designed a attention mechanism that combines the intermediate sentence feature representation from BILSTM and the local feature representation from CNN to generate a more robust sentence representation. Alshubaily\cite{Alshubaily2021} merged an attention mechanism into network to improve its performance. All this attention mechanisms are focus on word- or feature-level representation.

In our work, we don’t focus on keywords in a sentence, but which feature map of CNN is more important. We propose a Squeeze-and-Excitation Convolutional neural Network (SECNN) . SECNN takes the feature maps of multiple CNNs as different channels of sentence representation, and then, we can utilize channel attention mechanism, the SE attention mechanism, to enable the model to calculate the attention weights of different channels without any additional parameters. Our contributions are shown below:
\begin{itemize}
\item We innovatively introduce channel attention mechanism into sentence classification task;
\item We propose SECNN, in which the feature map from CNN is treated as a 'channel' of sentence representation, and channel attention mechanism can be utilized to re-weight these channels without any additional parameters;
\item Experiments show that the proposed model achieves advanced performance in the sentence classification task.
\end{itemize}

\section{SECNN}
\label{sec:unimodal}

The architecture of Squeeze-and-Excitation Convolutional Neural Network (SECNN) is presented in Fig. 1. The details of each component are as follows:

\begin{figure}[htbp]
\centering
\includegraphics[width=1\linewidth]{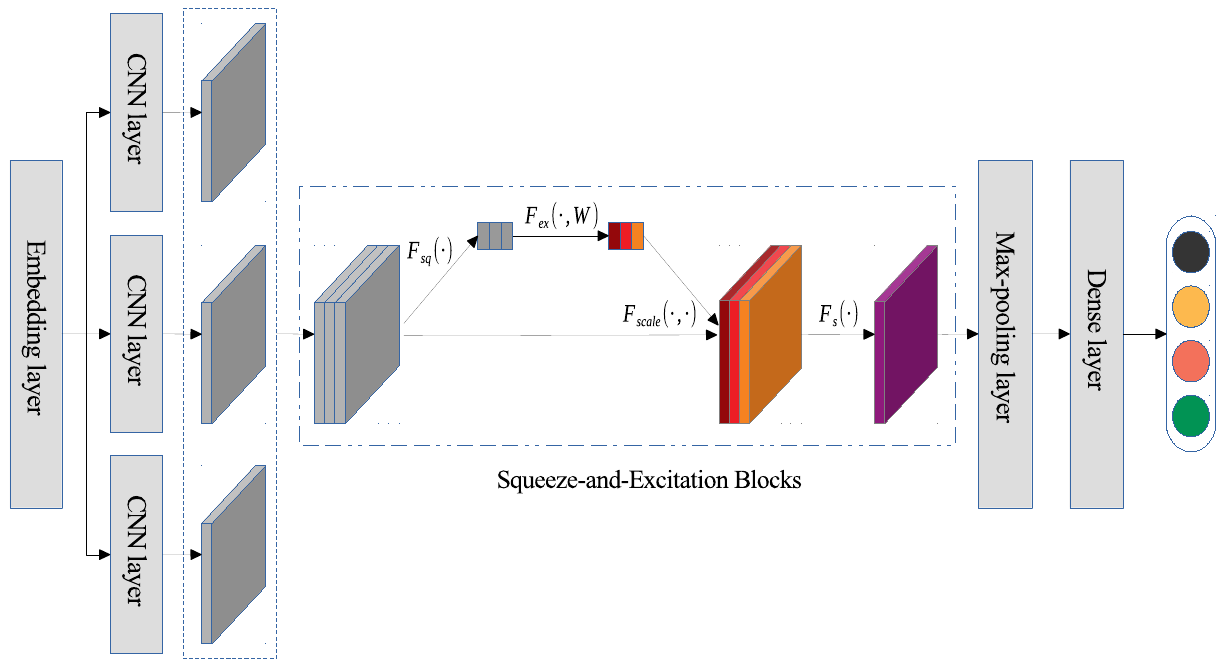}
\caption{The architecture of the proposed model}
\label{fig:1}
\end{figure}

\label{sec:methodology}
\subsection{Embedding Layer}
\label{sec:text}

The embedding layer converts every word $x_i$ in the sentence $s$ into a real-valued dense vector $ e_i $ whose dimension is $ d $. Thus, the sentence $s=[x_1, x_2, \cdots, x_n]$ of $n$ words can be represented as real-valued vector $e=[e_1, e_2, \cdots, x_n]$, $e \in \mathbb {R}^{n \times d}$. Here, we can use the randomly initialized embedding layer which will be modified during training, or the publicly available pre-trained Word2vec/Glove vectors which keep static.

\subsection{CNN Blocks}

The one-dimensional convolution is an operation that sliding a filter vector over a sequence to capture features at different position. For each input vector ${s} \in \mathbb{R}^{n \times d}$ and a filter vector ${f} \in \mathbb{R}^{k \times d}$, for each position $j$ of $s$, we have a window vector $w_j$ denoted as: $w_j=[x_j, x_{j+1}, \cdots, x_{j+k-1}]$ . Then, the convolution operation, that is, the filter f convolutes with the window vector $w_j$, Every CNN generates a feature map $c \in \mathbb{R}^{(n-k+1) \times d}$, $k$ is the length of filter. 

In order to obtain multiple feature maps of the same size, we use multiple filters of the same length. For $m$ filters, the generated $m$ feature maps can be stacked as $C=[c_1, c_2, \cdots, c_m]$ and $C \in \mathbb{R}^{H \times W \times M}$, here, $H=n-k+1$, $W=d$, $M=m$, $M$ is regarded as the channels of sentence representation.

\subsection{Squeeze-and-Excitation Blocks}
\label{sec:font}

Squeeze-and-Excitation Blocks (SE Blocks) is first proposed by Jie\cite{Jie2020} for vision tasks. Here, we utilize a variant of SE Blocks, the structure is depicted in Fig. 2.

\begin{figure}[htbp]
\centering
\includegraphics[width=0.7\linewidth]{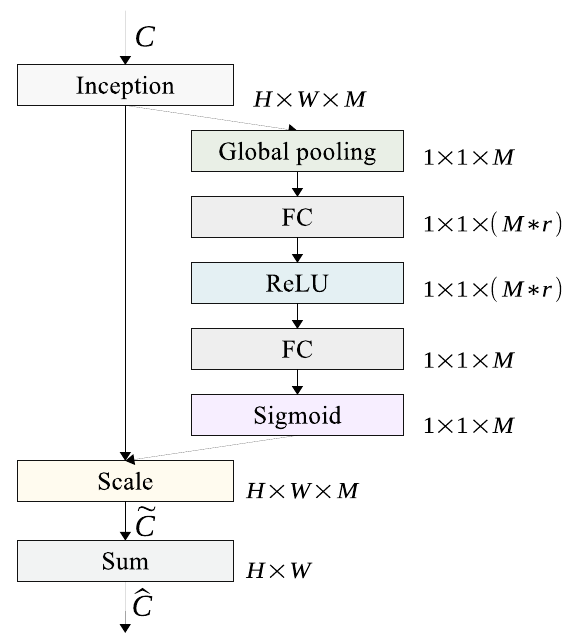}
\caption{The schema of Squeeze-and-Excitation Blocks}
\label{Fig:2}
\end{figure}

In order to obtain the weights of different channels, it squeezes global spatial information into a channel descriptor by using global average pooling to generate $channel$-wise statistic. Formally, the statistic $z \in \mathbb{R}^M$ is generated by shrinking $C$ through its spatial dimensions $H \times W$, such that the $m$-th element of $z$ is calculated by:

\begin{equation}
\label{eq1} z_m=F_{sq}(c_m)=\frac{1}{H \times W} \sum\limits_{i=1}^H \sum\limits_{j=1}^W C_m(i,j),
\end{equation}

The output of transformation C can be interpreted as a collection of channel descriptors whose statistics represent the entire sentence. Then, we employ a simple gating mechanism with activation functions to capture $channel$-wise dependencies $s$.

\begin{equation}
\label{eq2} s=F_{ex}(W,z)= \sigma(g(W,z))=\sigma(W_2\delta(W_1,z)),
\end{equation}

Here, $\sigma$ refers to the $sigmoid$, and $\delta$ refers to the $ReLU$, $W_1 \in \mathbb{R}^{(M*r) \times M}$, $W_2 \in \mathbb{R}^{M \times (M*r)}$ and $s=[s_1, s_2, \cdots, s_m]$.

And, two fully-connected layers are adopted around non-linearity to limit the complexity of the model and enhance generalization, i.e. a dimensionality-increasing layer with increasing ratio $r$, and then a dimensionality-reduction layer returning to the channel dimension of $C$. 

Then, scaling $C$ with $s$, the result can be obtained:

\begin{equation}
\label{eq3} \widetilde{C}=F_{scale}(C,s)=[c_1s_1, c_2s_2, \cdots, c_ms_m],
\end{equation}

Here, $\widetilde{C}=[\tilde{c_1}, \tilde{c_2}, \cdots, \tilde{c_m}]$ and, $F_{scale}$ refers to $channel$-wise multiplication between the scalar $s$ and the feature maps $C$.

The final output of SE block is obtained:

\begin{equation}
\label{eq4} \widehat{C}=F_s(\widetilde{C})=\sum\limits_{m=0}^M\tilde{c_m},
\end{equation}

Here, $\widehat{C} \in \mathbb{R}^{H \times W}$, and $F_s$ refers to $element$-wise add all $\tilde{c_m}$. 

\subsection{Classification Layer}

The outputs of SE block is passed to a $piece$-wise max-pooling layer which followed by dropout layer, and then, the output of the dropout layer is fed into a dense layer with $sigmoid$/$softmax$ activation to predict the label of a sentence in the given classification task.

\section{Experiments}
\subsection{Datasets}

We conducted experiments on several public datasets, the statistics of these datasets are as follows:
\begin{itemize}
\item MR. Movie Review—a movie review sentiment classification corpus contains 10662 sentences annotated with 2 labels;
\item IMDb. Internet Movie Database—a sentiment classification corpus contains 25000 sentences annotated with 2 labels;
\item AGNews. AG’s News Topic Classification Dataset—a news classification corpus contains 127600 sentences annotated with 4 labels;
\item DBpedia. A large-scale, multilingual knowledge base extracted from wikipedia—a topic classification corpus contains 630000 sentences annotated with 14 labels. 
\end{itemize}

\subsection{Hyperparameters and Training}

For MR, each sentence is padded to a maximum length of 50. For IMDb, each sentence is padded to a maximum length of 500. For AGNews, each sentence is padded to maximum length of 80. For DBPedia, each sentence is padded to a maximum length of 100. For all the datasets, the longer sentences are truncated.
For the CNN block, three parallel CNN layers with the same filter size of 3, with 128 feature maps each. dropout rate is 0.5, increasing ratio r is 16, piece of max-pooling is 3 and batch size of 64.
For all datasets, we randomly select 10$\%$ of the all samples as the dev set. 

\subsection{Results and Analysis}

The results of the proposed model against other models are listed in Table 1.

\doublerulesep 0.1pt
\begin{table}[h]
\begin{footnotesize}
\caption{Results of SECNN against other models} \label{tab:1}
\begin{tabular}{p{1.6cm}p{1.2cm}p{1.2cm}p{1.2cm}p{1.2cm}}
\hline\hline\noalign{\smallskip}
    Model& MR& IMDb& AGNews& DBpedia \\
\noalign{\smallskip} \hline
    CNN-rand\cite{Kim2014} & 0.7746 & 0.8986 & 0.8989 & 0.9850 \\
    C-LSTM\cite{Kalchbrenner2014}   & 0.7743 & 0.9009 & \pmb{0.9021} & 0.9845 \\
    ATT-CNN\cite{Zhou2015}  & \pmb{0.7844} & \pmb{0.9054} & 0.9015 & \pmb{0.9852} \\
    SECNN       & 0.7788 & 0.8870 & 0.8994 & 0.9827 \\
\hline\hline
\end{tabular}
\end{footnotesize}
\end{table}

The results show C-LSTM performs well on AGNews, ATT-CNN outperforms the others on MR, IMDb and DBpedia, while SECNN does not perform very well. We review the previous work, in SECNN, three parallel CNNs with same filter size of 3 are utilized to generate the same shape feature maps. So we speculate that the same filter size may reduce the diversity of the feature maps. Thus, we modify SECNN by utilizing three parallel CNNs with different filter sizes of 3, 4, and 5 to generate feature maps of the same size through preserving the convolution results at the boundaries. The performance of the method is shown in Table 2.

\doublerulesep 0.1pt
\begin{table}[h]
\begin{footnotesize}
\caption{Performances of the variant of SECNN} \label{tab:2}
\begin{tabular}{p{2.5cm}p{1.0cm}p{1.0cm}p{1.0cm}p{1.0cm}}
\hline\hline\noalign{\smallskip}s
    Model& MR& IMDb& AGNews& DBpedia \\
\noalign{\smallskip} \hline
    SECNN           & \underline{0.7862} & 0.8956 & 0.9017 & \underline{0.9858} \\
    SECNN-Glove     & 0.8134 & 0.9075 & \pmb{0.9209} & 0.9885 \\
    SECNN-Word2vec  & \pmb{0.8275} & \pmb{0.9137} & 0.9170 & \pmb{0.9890} \\
\hline\hline
\end{tabular}
\end{footnotesize}
\end{table}

The result show the variant of SECNN outperforms several superior baseline methods on two of the four datasets (MR and DBpedia). Besides, the pre-trained vectors with dimensionality of 300 from Word2vec and Glove are tested in our model, both of which are untrainable, static. The results show both Glove and Word2vec word vectors can improve performance. Among them, Glove performs better on AGNews, and Word2vec performs well on MR, IMDb and DBpedia.

In subsection 2.3, we introduce the increasing ratio $r$, a hyper-parameter. So we conduct experiment based on SECNN-Word2vec, the effects of increasing ratio on all datasets are presented in Fig.3.

\begin{figure}[!ht]
\centering
\includegraphics[width=1\linewidth]{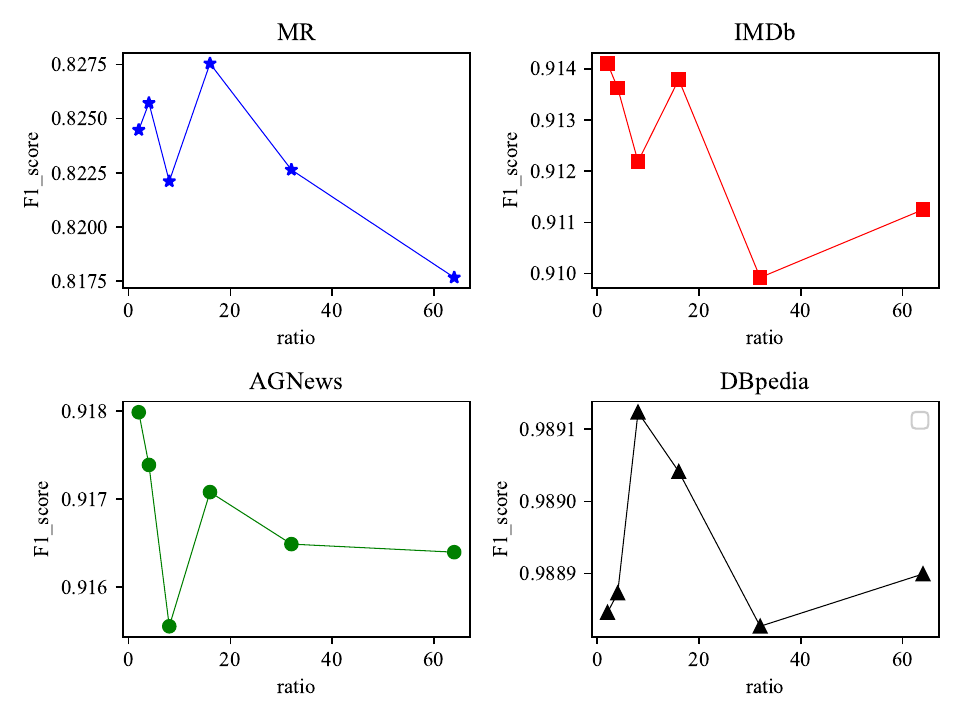}
\caption{Effects of increasing ratio}
\label{Fig3}
\end{figure}

Clearly, increasing ratio $r=\{2, 4,8,16,32,64\}$ behaves differently across the four datasets. Considering global performance, The proposed model sets increasing ratio r to 16. In practice, it may not be appropriate to use a same increasing ratio throughout the network, therefore, tuning the increasing ratio is necessary to meet the needs of a given model. 

\section{Conclusions and Future Work}

In our paper, we propose SECNN for sentence classification. The model introduces the SE attention mechanism to re-weight the feature maps of multiple CNNs without any additional parameters. Experiments show that SECNN outperforms several superior baseline methods on two of the four datasets. In future work, we will try to apply SECNN to other NLP tasks, and focus on exploring new attention mechanisms to compute attention weights on the feature representation of  multichannel layers (CNN or others).

\clearpage
\bibliographystyle{unsrt}
\bibliography{bibliography.bib}

\end{document}